\documentclass[]{spie}  

\usepackage{amsmath,amsfonts,amssymb}
\usepackage{graphicx}
\usepackage{float}
\usepackage{booktabs}
\usepackage{array}
\usepackage{enumitem}
\usepackage{microtype}
\usepackage{tikz}
\usetikzlibrary{arrows.meta,positioning,shapes.geometric}
\usepackage[colorlinks=true,allcolors=blue]{hyperref}
\hypersetup{
  pdftitle={A forensic evaluation framework for egocentric foundation models: an EFM3D case study for crime scene reconstruction},
  pdfauthor={Davidson Chisom Onwubiko; Shahana Bano; Carlos Francisco Moreno-García}
}
\setlist[enumerate]{nosep,leftmargin=*}

\title{A forensic evaluation framework for egocentric foundation models: an EFM3D case study for crime scene reconstruction}

\author{Davidson Chisom Onwubiko\supit{a}, Shahana Bano\supit{a}, and Carlos Francisco Moreno-Garc\'ia\supit{a}\skiplinehalf
\supit{a}School of Computing, Engineering \& Technology, Robert Gordon University, Aberdeen, United Kingdom}
\authorinfo{Author ORCID identifiers: Davidson Chisom Onwubiko, 0000-0002-7426-0126; Shahana Bano, 0000-0002-6907-1777; Carlos Francisco Moreno-Garc\'ia, 0000-0001-7218-9023.}

\begin{document}
\maketitle

\begin{abstract}
Foundation models increasingly combine geometry, semantics, and spatial context for three-dimensional scene understanding, but their suitability for forensic crime scene reconstruction remains unclear. This study presents a structured forensic evaluation framework for egocentric foundation models and demonstrates it using the Egocentric Foundation Model for 3D (EFM3D). Four simulated indoor crime scenes were recorded with Meta Project Aria Gen~1 at Robert 
Gordon University’s East Lodge House forensic training facility. The framework evaluates room-scale reconstruction, semantic scene understanding, computational performance, and forensic suitability. Across the four scenes, EFM3D generated meshes averaging $12{,}668\pm2{,}430$ vertices and $24{,}838\pm4{,}903$ faces, with 39,529 recorded semantic predictions spanning 21--24 categories at a mean confidence of 0.40. Because reference meshes and expert-labelled 3D ground truth were unavailable, these statistics are treated as descriptive outputs rather than measures of geometric or detection accuracy. Final inference runs processed $1{,}388\pm203$ snippets per scene at $0.69\pm0.01$ samples/s and required $50.77\pm6.12$ minutes per recording. A seven-criterion capability profile indicated stronger support for room geometry, furniture localisation, scene context, and spatial relationships than for fine-scale forensic evidence. The results position EFM3D as a complementary tool for contextual documentation and scene review rather than a replacement for high-fidelity forensic capture. The framework provides a reproducible basis for evaluating future egocentric foundation models in specialised forensic settings.
\end{abstract}


\section{INTRODUCTION}
Foundation models are reshaping machine vision by bringing geometry, semantics, and spatial reasoning into increasingly unified representations. In contrast to conventional pipelines that treat reconstruction, localisation, and object recognition as separate stages, recent multimodal models can reason jointly about the structure and content of a scene. This shift is particularly relevant in applications where a useful representation must capture not only where objects are located, but also what they are and how they relate to the surrounding environment.

Digital forensic investigation is one such application. Crime scene reconstruction preserves spatial relationships after a physical scene has been released and supports later examination, communication, training, and courtroom presentation \cite{ref1}. Photography, sketches, manual measurements, and total-station surveys remain important because they are established and defensible, yet they often leave investigators with fragmented records rather than a single explorable representation \cite{ref1,ref2,ref3,ref22}. Three-dimensional capture has improved this process: photogrammetry, terrestrial laser scanning (TLS), and LiDAR can preserve geometry with high accuracy \cite{ref3,ref4,ref5,ref6}, while Neural Radiance Fields (NeRF) and 3D Gaussian Splatting (3DGS) can produce photorealistic scene representations and flexible novel views \cite{ref7,ref8,ref13}. Even so, these approaches remain primarily concerned with geometry and appearance. Semantic interpretation identifying objects and understanding their contextual role is commonly added later or performed manually \cite{ref9,ref10} and remains an active area of methodological research in its own right, particularly with respect to loss-function design for segmentation accuracy \cite{ref19}.

Wearable sensing offers a different acquisition model. Meta Project Aria records synchronised visual, inertial, calibration, and other sensor streams from the wearer's perspective \cite{ref11}. Aria Machine Perception Services (MPS) convert these observations into trajectories, visual inertial odometry, and semi-dense geometry. EFM3D builds on this information through Egocentric Voxel Lifting (EVL), jointly predicting occupancy, object locations, and semantic context in three dimensions \cite{ref12}. In principle, a natural investigator walkthrough can therefore support both scene reconstruction and contextual interpretation.

The unresolved question is whether capabilities learned from general indoor environments transfer to forensic scenes. EFM3D was developed and evaluated primarily on everyday activities, whereas crime scenes may contain disturbed layouts, restricted viewpoints, severe occlusion, reflective surfaces, and small evidence whose position and condition must be preserved accurately \cite{ref1,ref12}. A model can therefore perform well at room-scale understanding while still failing on the details that matter most during evidential examination. Conventional machine-vision benchmarks do not fully capture this distinction. From a digital-forensics perspective, the resulting egocentric recordings, reconstructed geometry, semantic predictions, and spatial metadata form digital representations that can support subsequent scene examination, visualisation, and investigative review.

To address this gap, the present study introduces a structured forensic evaluation framework for egocentric foundation models and demonstrates it through an EFM3D case study using four simulated indoor crime scenes. The framework considers four complementary dimensions: room-scale reconstruction, semantic scene understanding, computational performance, and forensic suitability. Its purpose is not to replace established benchmark metrics, but to extend them with criteria that reflect practical forensic use.

The contribution of this work lies not in modifying EFM3D's architecture, but in introducing and demonstrating a forensic-domain evaluation methodology that separates conventional model outputs from the requirements of forensic use. The main contributions are:
\begin{enumerate}
\item a structured forensic evaluation framework for assessing egocentric foundation models in crime scene reconstruction;
\item an EFM3D case study based on a four-room Project Aria dataset collected within a dedicated forensic training facility;
\item a domain-specific evaluation methodology that distinguishes descriptive model outputs from forensic accuracy and operational readiness; and
\item evidence-based recommendations for forensic datasets, evidence-aware semantics, hybrid reconstruction, and future operational validation.
\end{enumerate}

The remainder of the paper reviews relevant reconstruction and egocentric-vision research, identifies the forensic evaluation gap, presents the framework and experimental methodology, reports the results, and discusses their implications and limitations.

\section{Related Work and Research Gap}
\subsection{Digital Crime Scene Reconstruction}
Digital reconstruction allows investigators to preserve spatial information after a physical scene has been released and supports measurement, review, and communication \cite{ref1,ref2}. Conventional documentation remains indispensable, but its fragmented nature can make complex scenes difficult to revisit and interpret as a whole \cite{ref2,ref3}. Photogrammetry, TLS, and LiDAR address this limitation by producing metrically useful 3D models~\cite{ref4,ref5,ref6}, while recent forensic
platforms have begun integrating 3D reconstruction with analytical tools for scene and trajectory interpretation~\cite{ref14}. NeRF and 3DGS further improve visual realism and novel-view rendering~\cite{ref7,ref8,ref13,ref15}, with recent work extending 3DGS toward photorealistic immersive VR applications \cite{ref18}.

These techniques differ in acquisition requirements and fidelity, yet most treat semantic understanding as a separate problem. A reconstruction may show where an object is located without identifying what it is or whether it has investigative relevance. Scene-labelling approaches can add semantics \cite{ref10}, but forensic interpretation still depends heavily on expert review \cite{ref9}. This creates a clear opportunity for systems that combine geometry with object-level reasoning.

\subsection{Egocentric Wearable Sensing and Foundation Models}
Egocentric devices capture a continuous first-person record as the wearer moves naturally through an environment. Project Aria combines RGB and monochrome scene cameras, inertial measurement units, eye cameras, microphones, and calibration data in a synchronised research platform \cite{ref11}. MPS estimates trajectories and semi-dense geometry, providing the spatial basis for downstream perception.

EFM3D extends this pipeline toward learned 3D scene understanding. Its EVL component lifts image features into a shared voxel representation from which the model predicts occupancy, oriented 3D bounding boxes, and semantic categories \cite{ref12}. The original EFM3D evaluation used Aria Digital Twin (ADT) and Aria Everyday Activities (AEA), which capture furnished apartments, offices, and ordinary activities rather than staged forensic evidence \cite{ref16,ref17}. This domain difference matters because models trained on everyday environments are naturally better represented for rooms, furniture, and common objects than for bloodstains, fingerprints, cartridge cases, fibres, or other evidence classes. Recent work on multimodal and geospatial forensic reconstruction further illustrates the need to preserve evidence-specific detail alongside scene-level geometry \cite{ref22,ref23}.

\subsection{Research Gap}
Existing evaluations of egocentric foundation models emphasise reconstruction, detection, segmentation, and scene understanding within general indoor environments. Those benchmarks are valuable for measuring technical progress, but they do not ask whether outputs preserve evidential detail, provide meaningful forensic context, or remain practical within an investigative workflow. Forensic suitability therefore cannot be inferred from benchmark performance alone.

To our knowledge, structured evaluation of EFM3D under these forensic requirements remains limited. The present work addresses that gap by combining conventional machine-vision measurements with an explicit forensic suitability assessment. The resulting framework is intended as a reproducible evaluation methodology rather than a claim of universal forensic validation.

\section{Forensic Evaluation Framework and Experimental Methodology}\label{sec:experimental-design}
\subsection{Framework Motivation and Overview}
A forensic reconstruction system must do more than recover visually plausible geometry. It should preserve the spatial arrangement of the scene, provide semantically useful context, operate within a practical processing workflow, and avoid overstating evidential detail that it cannot reliably represent. These requirements motivate a broader evaluation than geometry-only or detection-only benchmarks.

\begin{figure}[H]
\centering
\includegraphics[width=0.75\linewidth]{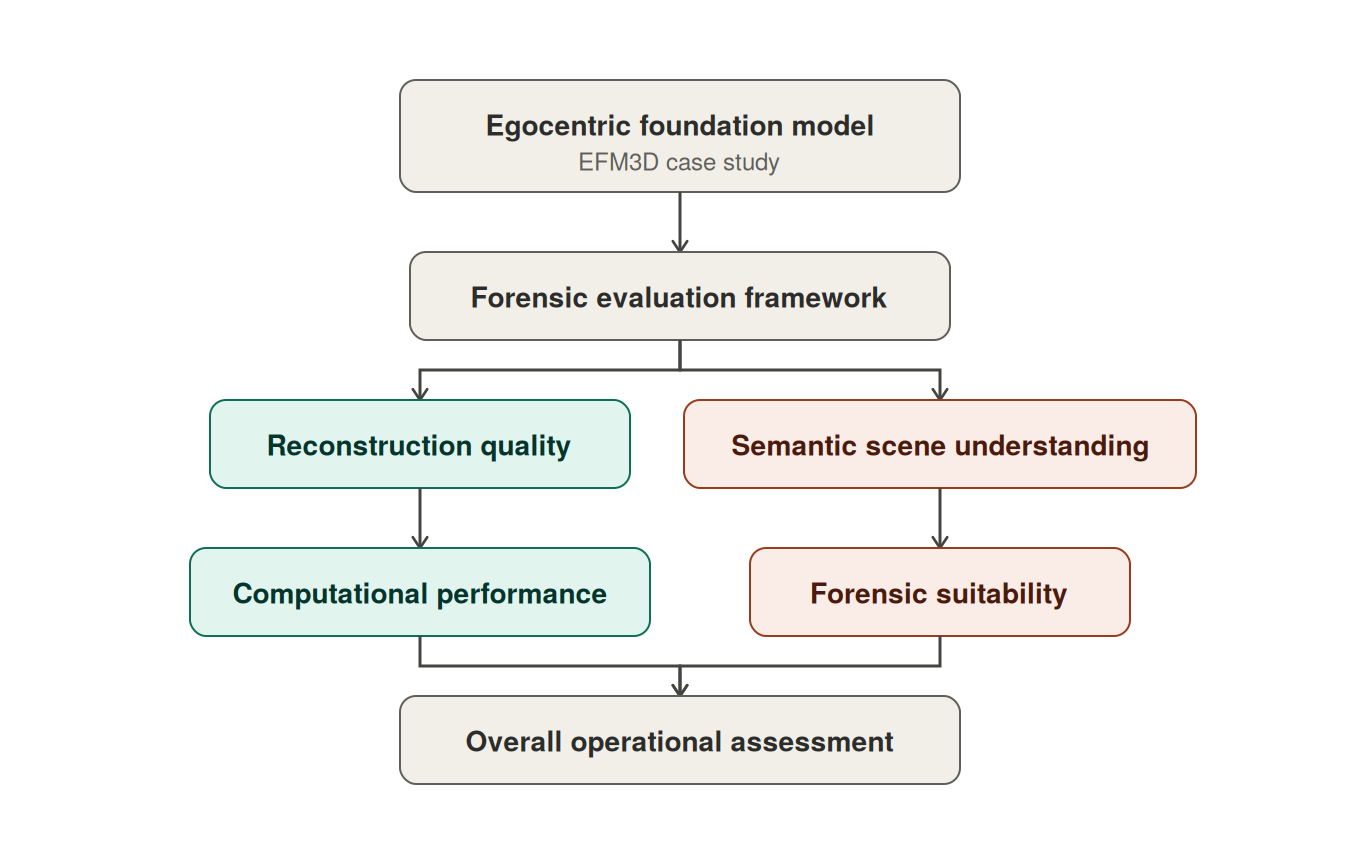}
\caption{Proposed forensic evaluation framework for assessing egocentric foundation models. The framework evaluates an egocentric foundation model through four complementary dimensions: room-scale reconstruction, semantic scene understanding, computational performance, and forensic suitability, which together provide an overall operational assessment. In this study, the framework is demonstrated using EFM3D as the case study.}
\label{fig:framework}
\end{figure}

The proposed framework therefore organises assessment around four dimensions: \textbf{room-scale reconstruction}, \textbf{semantic scene understanding}, \textbf{computational performance}, and \textbf{forensic suitability}. Figure~\ref{fig:framework} shows how these dimensions connect the model outputs to an overall operational assessment. The framework is deliberately model-agnostic: EFM3D is used here as the case study, but the same dimensions can support future evaluations of other egocentric foundation models.

\subsection{Objectives and Research Questions}
The framework is demonstrated using the official EFM3D pipeline and four simulated indoor crime scenes. Table~\ref{tab:domain} contrasts the forensic training-facility data with the principal datasets originally used to evaluate EFM3D.

\begin{table}[t]
\caption{Domain comparison between EFM3D benchmarks and the present dataset.}
\label{tab:domain}
\centering
\scriptsize
\setlength{\tabcolsep}{2.2pt}
\renewcommand{\arraystretch}{1.03}
\begin{tabular}{p{0.14\textwidth}p{0.25\textwidth}p{0.22\textwidth}p{0.31\textwidth}}
\toprule
\textbf{Characteristic} & \textbf{ADT \cite{ref16}} & \textbf{AEA \cite{ref17}} & \textbf{Forensic training facility} \\
\midrule
Environment & Furnished apartment and office & Everyday activities in real homes & Four staged indoor crime-scene rooms \\
Sequences & 200 & 143 & 4 \\
Volume & Sequence count reported & 7.3 h; $>$1M frames & 563.7 s; 16,921 RGB frames \\
Object ground truth & 398 instances with 6DoF poses and segmentation & None published & No manual ground truth; EFM3D outputs only \\
Semantic scope & Everyday household/office objects & No released semantic labels & 21--24 predicted everyday categories; no forensic classes \\
Scene intent & Natural domestic/office activity & Natural daily activity & Disturbed layouts and forensic walkthroughs \\
Evaluation focus & Detection, segmentation, reconstruction, pose & Long-form multimodal egocentric AI & Reconstruction, semantics, performance, and forensic suitability \\
\bottomrule
\end{tabular}
\end{table}

The four research questions correspond directly to the framework dimensions:
\begin{description}
\item[RQ1] What room-scale reconstruction characteristics does EFM3D exhibit within simulated forensic environments?
\item[RQ2] What semantic object predictions does EFM3D produce in simulated forensic environments, and what limitations are evident for forensic scene understanding?
\item[RQ3] What computational characteristics are observed when processing Project Aria recordings?
\item[RQ4] To what extent are the generated outputs suitable for forensic crime scene documentation and interpretation?
\end{description}

\subsection{Crime Scene Dataset and Acquisition}
Data were collected at East Lodge House, Robert Gordon University’s forensic training facility. Four furnished rooms with different layouts and object distributions were treated as independent simulated crime scenes. The available dataset records room-level capture characteristics but does not contain a formal scene-by-scene inventory of staged evidence; consequently, no additional scene composition or evidence-count claims are made here. Each recording followed the same broad protocol: the wearer moved naturally around the room, covered the principal viewpoints, and avoided abrupt head movements that could disrupt visual--inertial tracking.

\begin{figure}[H]
\centering
\includegraphics[width=0.32\textwidth]{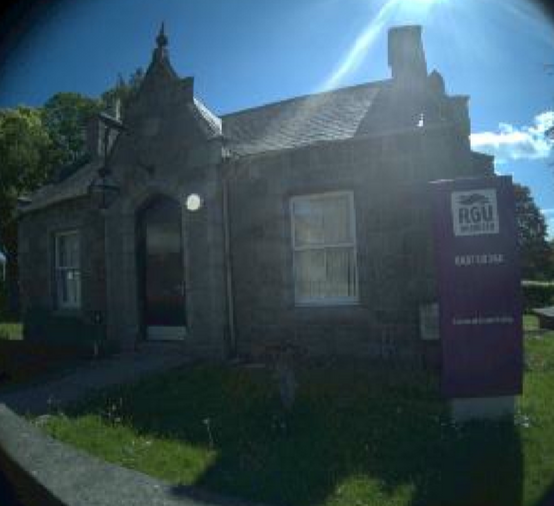}
\caption{East Lodge House forensic training facility used for data collection.}
\label{fig:facility}
\end{figure}

Project Aria Gen~1 was operated with Recording Profile~28, which provides the synchronised streams required by MPS and EFM3D \cite{ref11,ref12}. The four VRS recordings totalled 563.69~s, 16,921 RGB frames, and 8.642~GB. Their characteristics are reported in Table~\ref{tab:dataset}; the sensor profile is summarised in Table~\ref{tab:profile}.

\begin{table}[H]
\caption{Forensic training facility evaluation dataset.}
\label{tab:dataset}
\centering
\small
\setlength{\tabcolsep}{3pt}
\renewcommand{\arraystretch}{1.02}
\resizebox{\textwidth}{!}{%
\begin{tabular}{lccccc}
\toprule
\textbf{Metric} & \textbf{Room 1} & \textbf{Room 2} & \textbf{Room 3} & \textbf{Room 4} & \textbf{Mean} \\
\midrule
Duration (s) & 112.61 & 157.54 & 153.39 & 140.15 & 140.92 \\
VRS size (GB) & 1.314 & 1.962 & 1.762 & 1.604 & 1.661 \\
RGB frames & 3,380 & 4,729 & 4,605 & 4,207 & 4,230 \\
RGB rate (fps) & 30.02 & 30.02 & 30.02 & 30.02 & 30.02 \\
\bottomrule
\end{tabular}%
}
\end{table}

\begin{table}[H]
\caption{Project Aria Recording Profile 28.}
\label{tab:profile}
\centering
\scriptsize
\setlength{\tabcolsep}{4pt}
\renewcommand{\arraystretch}{1.0}
\begin{tabular}{llll}
\toprule
\textbf{Stream} & \textbf{Resolution} & \textbf{Rate} & \textbf{Format} \\
\midrule
RGB camera & $1408\times1408$ px & 30 fps & JPEG \\
SLAM cameras ($\times2$) & $640\times480$ px & 30 fps & JPEG \\
Eye cameras ($\times2$) & $320\times240$ px & 60 fps & JPEG \\
IMU 1 / IMU 2 & --- & 1000 / 800 Hz & --- \\
Magnetometer / barometer & --- & 10 / 50 Hz & --- \\
Microphones ($\times7$) & --- & 48 kHz & --- \\
\bottomrule
\end{tabular}
\end{table}

\subsection{Experimental Platform}

The experiments used the official Meta EFM3D implementation without algorithmic modification. Processing was performed in Google Colaboratory on Ubuntu~24.04~LTS using Python~3.12, PyTorch~2.2.2, Open3D~0.19, CUDA~12.8, and an NVIDIA A100-SXM4-40GB GPU. Both official EFM3D checkpoints were included in the released checkpoint package; however, the inference workflow reported in this study used \texttt{model\_lite.pth}. Project Aria recordings were pre-processed using Aria Machine Perception Services (MPS) before EFM3D inference.

\begin{table}[H]
\caption{Experimental hardware and software configuration.}
\label{tab:platform}
\centering
\scriptsize
\setlength{\tabcolsep}{4pt}
\renewcommand{\arraystretch}{1.02}
\begin{tabular}{ll}
\toprule
\textbf{Component} & \textbf{Specification} \\
\midrule
Wearable sensor & Meta Project Aria Gen~1 \\
Recording profile & Profile 28 \\
Operating system & Ubuntu 24.04 LTS \\
Development environment & Google Colaboratory \\
Programming language & Python 3.12 \\
GPU & NVIDIA A100-SXM4-40GB \\
CUDA & 12.8 \\
Deep learning framework & PyTorch 2.2.2 \\
3D processing library & Open3D 0.19 \\
Scene understanding & EFM3D (official Meta implementation) \\
Inference checkpoint & \texttt{model\_lite.pth} \\
Perception pipeline & Aria Machine Perception Services (MPS) \\
\bottomrule
\end{tabular}
\end{table}

\subsection{Processing Pipeline}
\begin{figure}[H]
\centering
\includegraphics[width=\textwidth,height=0.38\textheight,keepaspectratio]{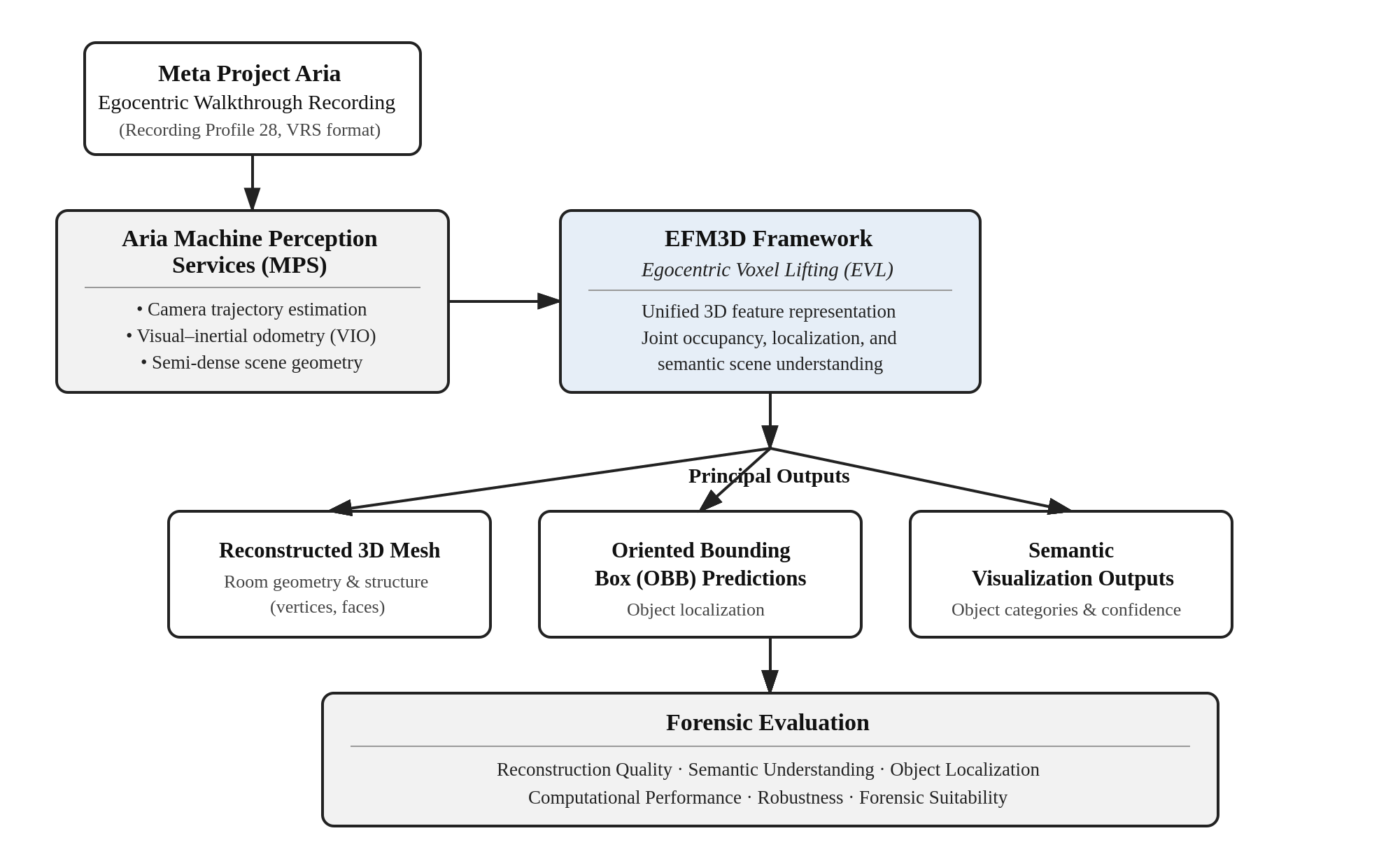}
\caption{Project Aria acquisition, MPS pre-processing, EFM3D inference, and forensic evaluation pipeline.}
\label{fig:workflow}
\end{figure}

The official Meta workflow was retained without architectural modification (Fig.~\ref{fig:workflow}). MPS first generated camera trajectories, visual--inertial odometry, and semi-dense geometry. EFM3D then applied EVL to form a 3D feature representation and produced a reconstructed mesh, oriented bounding boxes, and semantic visualisations \cite{ref11,ref12}. These outputs were subsequently assessed using the four dimensions of the proposed framework.

\subsection{Evaluation and Analysis}
Quantitative analysis used recording duration, frame count, mesh statistics, semantic predictions, confidence scores, processing logs, throughput, and inference time. Qualitative assessment was used where the dataset lacked manually annotated forensic ground truth. Because no reference meshes or expert-labelled 3D object annotations were available, mAP, IoU, F1-score, and absolute reconstruction error could not be calculated. The analysis therefore does not claim benchmark-level detection accuracy; instead, it characterises consistency across scenes and evaluates practical scene-documentation value.

For the forensic-suitability dimension, seven operational criteria were examined using a categorical capability profile. Each criterion was classified as supported, partially supported, limited, or not supported on the basis of the observed outputs. These categories are descriptive rather than numerical: they make the assessment criteria explicit and expose relative strengths and limitations within this case study without implying a validated forensic-readiness score. No validated weighting scheme or multi-rater instrument currently exists for this application. Future work should establish expert-derived criteria and inter-rater reliability before a formal forensic-readiness instrument is considered.

\section{Evaluation Results}\label{sec:results}
\subsection{RQ1: Room-Scale Reconstruction Characteristics}
The first framework dimension characterises room-scale reconstruction behaviour across scenes with different layouts and object distributions. Table~\ref{tab:reconstruction} reports descriptive reconstruction statistics. EFM3D generated a recognisable room-scale model for every recording, with meshes averaging $12{,}668\pm2{,}430$ vertices and $24{,}838\pm4{,}903$ faces. Room~3 produced the largest mesh and Room~1 the smallest. These counts describe output density and consistency only: without a reference mesh they cannot establish geometric accuracy, completeness, or superiority. The effective acquisition rate remained approximately 30~fps across all recordings, indicating stable capture.

\begin{table}[H]
\caption{Reconstruction characteristics across the four simulated crime scenes.}
\label{tab:reconstruction}
\centering
\small
\setlength{\tabcolsep}{2.6pt}
\renewcommand{\arraystretch}{1.02}
\resizebox{\textwidth}{!}{%
\begin{tabular}{lccccc}
\toprule
\textbf{Metric} & \textbf{R1} & \textbf{R2} & \textbf{R3} & \textbf{R4} & \textbf{Mean $\pm$ SD} \\
\midrule
Duration (s) & 112.61 & 157.54 & 153.39 & 140.15 & 140.92 $\pm$ 20.30 \\
RGB frames & 3,380 & 4,729 & 4,605 & 4,207 & 4,230 $\pm$ 608 \\
RGB rate (fps) & 30.02 & 30.02 & 30.02 & 30.02 & 30.02 $\pm$ 0.00 \\
Mesh vertices & 10,485 & 12,203 & 16,235 & 11,750 & 12,668 $\pm$ 2,430 \\
Mesh faces & 20,590 & 23,811 & 31,904 & 23,045 & 24,838 $\pm$ 4,903 \\
\bottomrule
\end{tabular}%
}
\end{table}

\begin{figure}[H]
\centering
\includegraphics[width=0.475\textwidth]{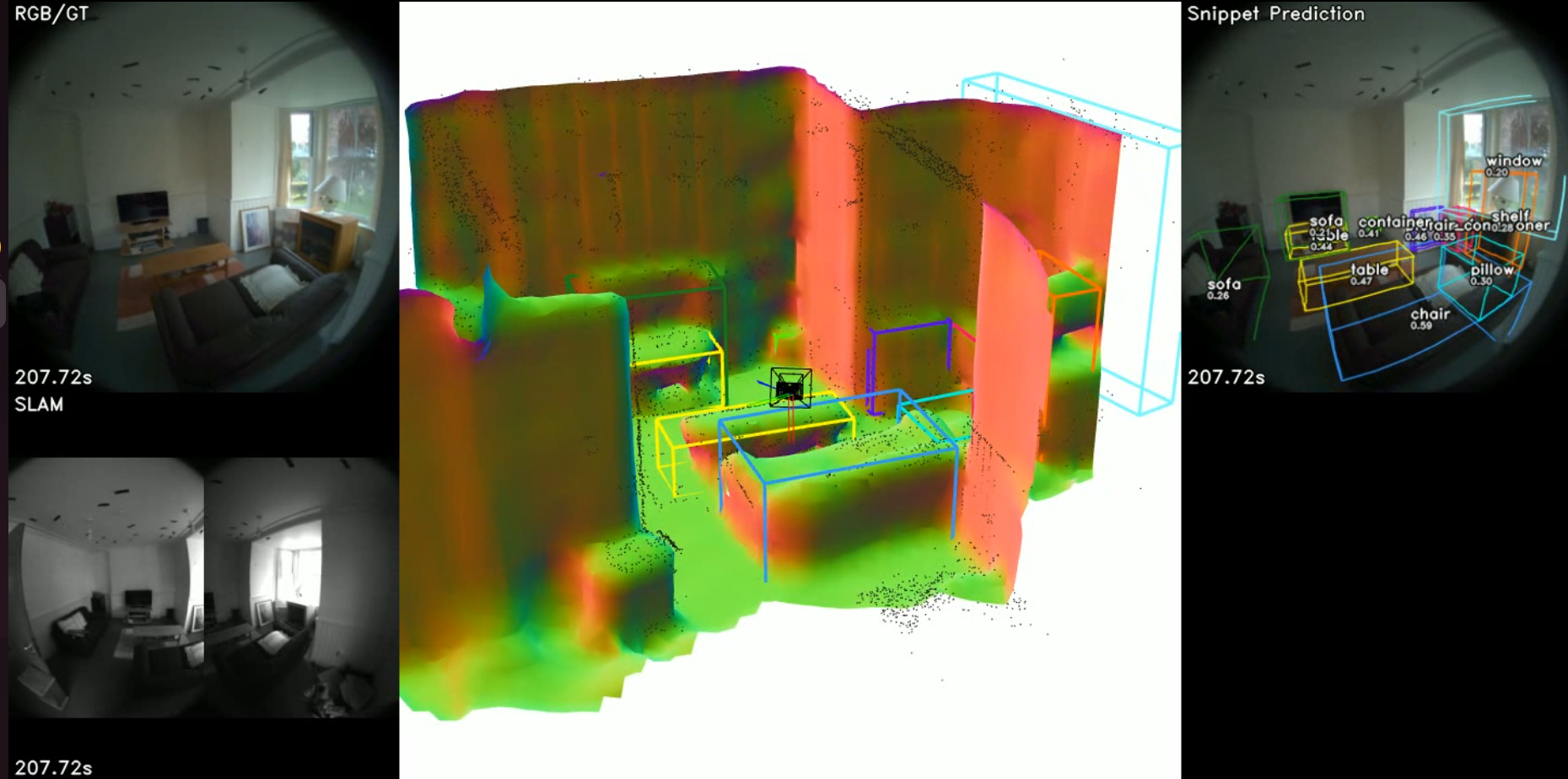}\hfill
\includegraphics[width=0.475\textwidth]{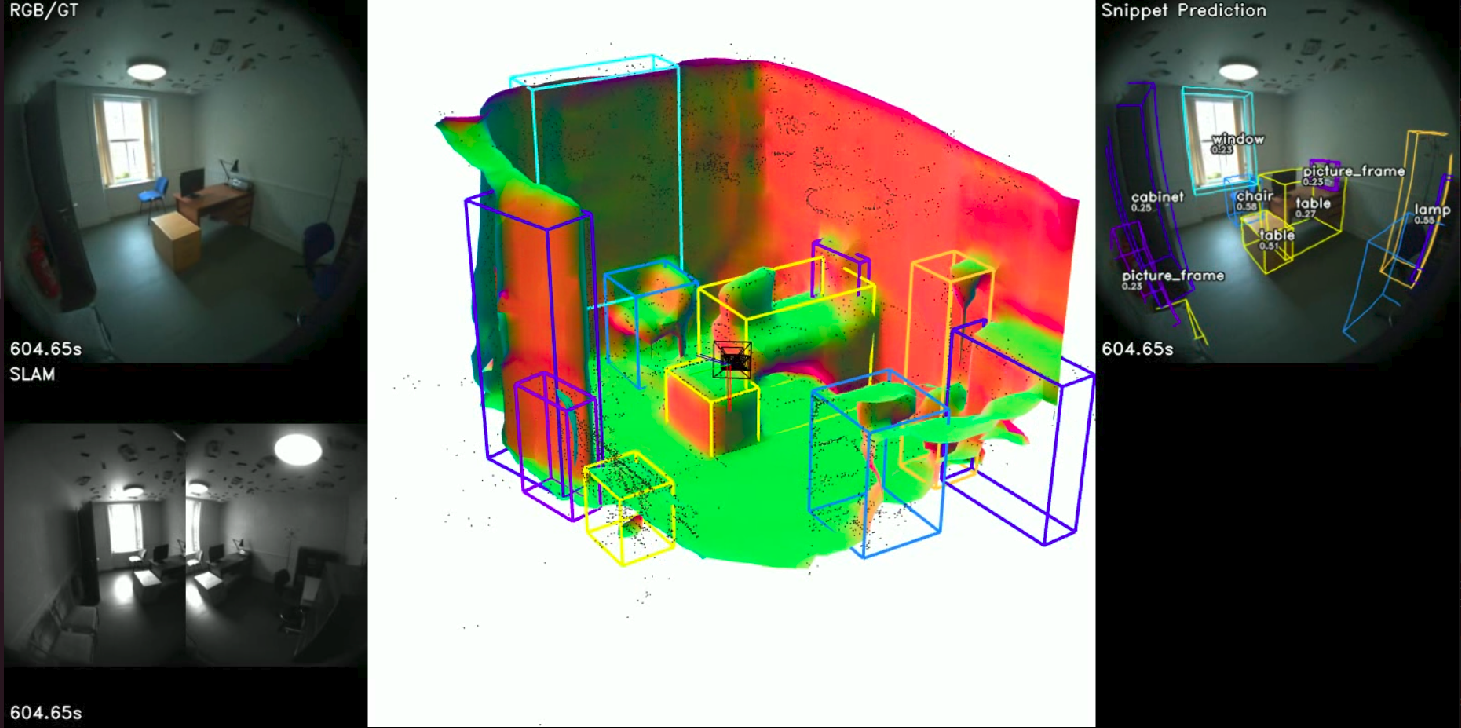}
\caption{Representative EFM3D outputs for Rooms 1 (left) and 2 (right).}
\label{fig:rooms12}
\end{figure}

\begin{figure}[H]
\centering
\includegraphics[width=0.475\textwidth]{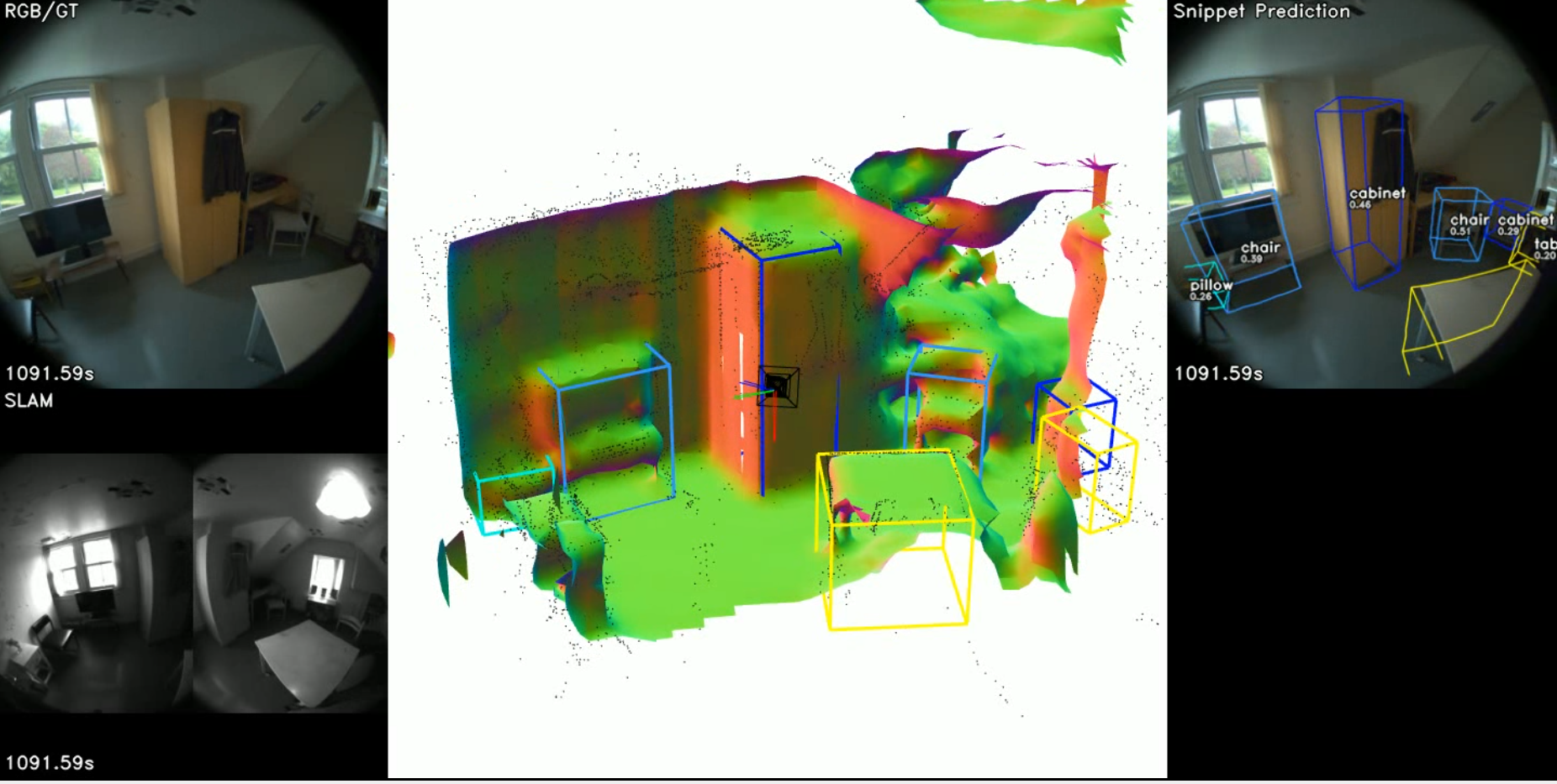}\hfill
\includegraphics[width=0.475\textwidth]{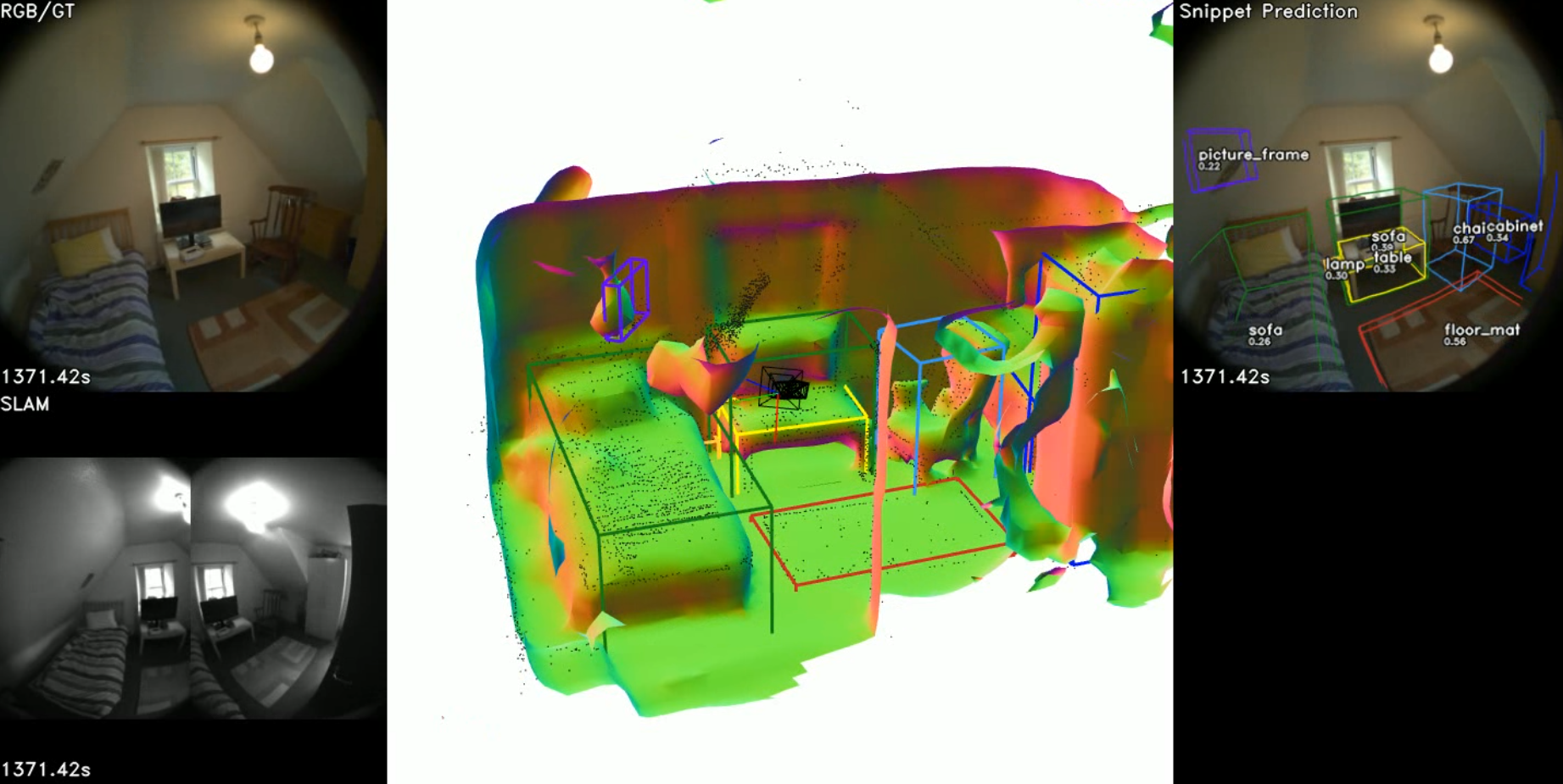}
\caption{Representative EFM3D outputs for Rooms 3 (left) and 4 (right).}
\label{fig:rooms34}
\end{figure}

Figures~\ref{fig:rooms12} and~\ref{fig:rooms34} show the corresponding RGB/SLAM views, reconstructed surfaces, oriented bounding boxes, and semantic predictions. Qualitative inspection showed visually coherent representations of the main room boundaries and furniture across the four scenes. Local surface discontinuities and noisy regions were nevertheless visible where coverage or visual texture was limited. Emerging point-cloud completion techniques, which infer and fill missing geometric detail from partial observations, may offer a route to addressing such reconstruction gaps in future work \cite{ref20}. These observations are consistent with sensitivity to scene coverage, acquisition trajectory, and available visual texture, although these factors were not independently manipulated in the present study.

\subsection{RQ2: Semantic Scene Understanding}
The second framework dimension characterises the semantic object predictions produced by EFM3D and their limitations for forensic scene understanding. Across the four scenes, EFM3D generated 39,529 recorded semantic predictions spanning 21--24 semantic categories (Table~\ref{tab:semantic}). Chairs, tables, cabinets, sofas, lamps, and other large indoor objects dominated the output.

\begin{table}[H]
\caption{Semantic prediction summary.}
\label{tab:semantic}
\centering
\small
\begin{tabular}{lr}
\toprule
\textbf{Metric} & \textbf{Value} \\
\midrule
Total recorded predictions & 39,529 \\
Mean predictions per scene & 9,882 \\
Categories per scene & 21--24 \\
Mean confidence & 0.400 \\
Mean predictions per RGB frame & 2.34 \\
Mean simultaneous predictions & 7.14 \\
Detection coverage & 32.80\% \\
\bottomrule
\end{tabular}
\end{table}

The semantic outputs also expose a clear domain limitation. Qualitative inspection indicated that small objects were less consistently represented, while forensic categories such as fingerprints, bloodstains, and trace material were absent from the model vocabulary. The mean confidence of 0.40 further indicates uncertainty in some everyday-object predictions. These observations suggest greater utility for broad contextual interpretation than for fine-grained evidential analysis. Detection coverage was approximately 32.8\% across all scenes; this value denotes the proportion of RGB frames with at least one recorded semantic detection and is not an intrinsic accuracy measure.

\subsection{RQ3: Computational Performance}
The third framework dimension examines whether the workflow is practical for offline forensic processing. Table~\ref{tab:compute} reports the final inference measurements. EFM3D processed between 1,105 and 1,554 snippets per scene at a stable throughput of 0.67--0.70 samples/s. Mean inference time was $50.77\pm6.12$ minutes per recording. Runtime increased with the amount of data processed, but throughput remained tightly clustered across the four scenes, indicating stable computational behaviour under the final experimental configuration.

\begin{table}[H]
\caption{Computational performance of EFM3D across the four recordings.}
\label{tab:compute}
\centering
\scriptsize
\setlength{\tabcolsep}{2.5pt}
\resizebox{\textwidth}{!}{%
\begin{tabular}{lccccc}
\toprule
\textbf{Metric} & \textbf{R1} & \textbf{R2} & \textbf{R3} & \textbf{R4} & \textbf{Mean $\pm$ SD} \\
\midrule
Snippets processed & 1,105 & 1,554 & 1,513 & 1,380 & 1,388 $\pm$ 203 \\
Throughput (samples/s) & 0.70 & 0.69 & 0.69 & 0.67 & 0.69 $\pm$ 0.01 \\
Inference time (s) & 2580.41 & 3408.28 & 3257.84 & 2938.72 & 3046.31 $\pm$ 367.15 \\
Inference time (min) & 43.01 & 56.80 & 54.30 & 48.98 & 50.77 $\pm$ 6.12 \\
VRS size (GB) & 1.314 & 1.962 & 1.762 & 1.604 & 1.661 $\pm$ 0.278 \\
\bottomrule
\end{tabular}%
}
\end{table}

The observed runtime is not suitable for real-time deployment, but real-time operation was not the objective of the present study. For the short recordings evaluated here, offline inference required approximately 43--57 minutes per scene. Whether this processing time is operationally acceptable in forensic practice requires practitioner-centred evaluation. The measurements should therefore be interpreted as hardware- and implementation-specific rather than as a general benchmark of EFM3D speed.

\subsection{RQ4: Forensic Suitability}
The fourth and final framework dimension relates the observed technical behaviour to practical forensic requirements. Table~\ref{tab:suitability} presents a seven-criterion categorical capability profile. Within this case study, room geometry, furniture localisation, and spatial relationships were classified as supported; scene context as partially supported; small-evidence representation as limited; and bloodstain and fingerprint recognition as not supported. These classifications constitute an exploratory single-assessor assessment rather than a validated multi-rater instrument.

\begin{table}[H]
\caption{Structured forensic suitability assessment using the categorical capability profile defined in Section~3.6.}
\label{tab:suitability}
\centering
\scriptsize
\setlength{\tabcolsep}{2pt}
\renewcommand{\arraystretch}{1.04}
\begin{tabular}{p{0.34\textwidth}p{0.22\textwidth}p{0.36\textwidth}}
\toprule
\textbf{Criterion} & \textbf{Capability} & \textbf{Basis in observed outputs} \\
\midrule
Room geometry reconstruction & Supported & Recognisable room-scale meshes in all four scenes \\
Furniture localisation & Supported & Common furniture predictions present across scenes \\
Scene context understanding & Partially supported & Broad indoor semantic context represented, with limitations \\
Spatial relationships & Supported & Room-scale geometry and object locations jointly represented \\
Small-evidence representation & Limited & Qualitative inspection indicated weak/inconsistent representation \\
Bloodstain recognition & Not supported & Forensic-specific class absent from model vocabulary \\
Fingerprint recognition & Not supported & Forensic-specific class absent from model vocabulary \\
\bottomrule
\end{tabular}
\end{table}

The profile exposes a scale-dependent boundary in the observed outputs: room layout, furniture, and broad spatial context were represented more effectively than trace-level evidence. The classifications should therefore be read as a transparent case-study synthesis rather than as a validated measure of forensic readiness.

\subsection{Relationship to Existing Reconstruction Approaches}
EFM3D addresses a different part of the reconstruction workflow from geometry-first techniques. Photogrammetry, TLS/LiDAR, NeRF, and 3DGS primarily preserve geometry or appearance, whereas EFM3D combines wearable egocentric acquisition with learned 3D occupancy and semantic object predictions \cite{ref2,ref3,ref4,ref5,ref6,ref7,ref8,ref12,ref13}. The methods were not benchmarked on a common reference scene in this study; consequently, no ranking of their geometry, speed, or evidence-preservation performance is made. The present results instead motivate a hybrid evaluation question: whether semantic egocentric scene understanding can complement established high-fidelity capture while preserving clear boundaries between model inference and evidential observation. Comparative experiments using common reference scenes and ground-truth geometry are required to test that proposition.

\subsection{Summary of Findings}
Application of the proposed framework shows that EFM3D produced interpretable room-scale reconstructions, common-object semantic predictions, and stable offline computational behaviour across the four simulated scenes. The framework also exposes an important domain gap: forensic-specific and fine-scale evidence was not adequately represented in the observed outputs. These findings support further investigation of egocentric foundation models as complementary tools for contextual documentation and scene review, while leaving evidential accuracy and operational readiness to be established through ground-truth and practitioner-centred validation.

\section{Discussion}
\subsection{Interpretation of the Findings}
The case study demonstrates that a general-purpose egocentric foundation model can transfer useful room-level knowledge to a forensic setting without architectural modification. Across all four recordings, EFM3D produced representations of the dominant indoor structure and predictions corresponding to common furniture, combining room-scale geometry with semantic context. These observed capabilities motivate further evaluation for tasks such as scene familiarisation, collaborative review, training, and preparation of immersive scene interfaces.

The results also show why good scene understanding should not be confused with evidential completeness. The observed pattern is consistent with EFM3D's general indoor-domain training and evaluation context: furniture and common indoor objects were more readily represented, whereas forensic-specific and trace-level evidence was absent or under-resolved. A coherent room-scale mesh can therefore coexist with the loss of precisely the details that make a reconstruction evidentially useful. The distinction is central to forensic deployment.

Wearable acquisition is nevertheless attractive. A natural walkthrough can preserve a continuous first-person record without occupying the investigator's hands or requiring a scanner to be repositioned at multiple stations. In time-sensitive settings, this may provide a rapid contextual record before targeted high-resolution evidence capture begins. The present findings therefore support investigating EFM3D as a complementary rather than substitutive component of forensic scene-documentation workflows.

\subsection{Implications for Machine Vision}
The findings have implications beyond digital forensics. Foundation models are commonly judged using general benchmarks that emphasise reconstruction, detection, segmentation, or localisation. Those measures remain essential, but they provide only a partial view of suitability once a system enters a specialised or safety-critical domain. Application-specific requirements can expose limitations that remain invisible in benchmark scores.

The proposed framework illustrates one way to extend evaluation without discarding conventional metrics. By combining technical measurements with explicit operational criteria, it separates model capability from domain readiness. Similar reasoning may be relevant wherever foundation models are expected to support decisions in environments where missing small details, misinterpreting context, or overstating confidence carries practical consequences.

\subsection{Implications for Digital Forensics}
For forensic practice, the immediate value of egocentric foundation models lies in contextual documentation rather than evidential replacement. EFM3D can help an examiner understand room layout, navigate a digital scene, and identify major objects, but high-fidelity photography, photogrammetry, TLS, LiDAR, and specialist evidence imaging remain necessary for precise measurement and detailed examination \cite{ref2,ref3,ref4,ref5,ref6}.

A hybrid workflow is therefore more realistic: wearable egocentric capture provides rapid scene-level context, while targeted high-resolution methods preserve evidentially important regions. Future systems could combine these complementary representations within a common spatial frame, allowing semantic scene understanding to guide investigators toward areas that require more detailed capture without presenting inferred semantics as evidence.

\subsection{Threats to Validity}
Several factors constrain the interpretation of the study. First, the evaluation used four simulated indoor crime scenes from a single training facility. The scenes provide controlled forensic context, but they do not represent outdoor, industrial, public, or highly dynamic environments. Second, the study evaluates a single egocentric foundation model using the official Meta implementation without task-specific fine-tuning on forensic datasets. Although EFM3D represents one of the most advanced publicly available egocentric foundation models, alternative models or domain-adapted versions may demonstrate different reconstruction characteristics and semantic capabilities.

Third, the dataset does not contain manually annotated 3D forensic ground truth. This prevented calculation of mAP, IoU, pose error, completeness, or metric reconstruction accuracy and required the study to combine objective output statistics with structured qualitative assessment. Fourth, the forensic suitability ratings were produced as an exploratory assessment rather than through multiple independent practitioners. Inter-rater reliability and alternative weighting strategies therefore remain to be validated. The recordings were also produced using a single acquisition procedure, so the effects of operator behaviour, viewing trajectory, walking speed, viewing distance, and scene coverage were not independently evaluated. Finally, the study addresses technical suitability rather than chain of custody, evidential provenance, explainability, repeatability, legal admissibility, or end-user effectiveness.

These limitations do not invalidate the case study, but they define its scope. The proposed framework should be regarded as a structured starting point for forensic evaluation rather than a validated standard.

\subsection{Directions for Future Work}
Future development should focus on four priorities. First, larger multi-site egocentric datasets are needed with expert labels for evidence type, 3D position, visibility, and uncertainty. Second, domain adaptation should introduce evidence-aware categories and contextual relationships rather than relying solely on everyday-object vocabularies. Third, comparative studies should apply the same framework to photogrammetry, LiDAR, NeRF, 3DGS, and emerging foundation models using common reference scenes, complemented by systematic comparison of object detection architectures suited to evidence-aware semantic categories \cite{ref21}. Finally, practitioner-centred validation should assess inter-rater reliability, uncertainty communication, provenance, and XR-based review with investigators \cite{ref9}.

\section{Conclusion}
This study presented a structured forensic evaluation framework for egocentric foundation models and demonstrated it through an EFM3D case study using four Project Aria recordings from simulated indoor crime scenes. The framework brings together room-scale reconstruction, semantic scene understanding, computational performance, and forensic suitability so that technical outputs can be interpreted in relation to operational forensic requirements.

The case study showed that EFM3D produced interpretable room-scale geometry and predictions for major furniture while adding scene-level semantic context. Inference behaviour was stable across the four recordings, with an average runtime of $50.77\pm6.12$ minutes under the final experimental configuration. At the same time, the evaluation exposed a clear evidential boundary: fine-scale evidence and forensic-specific semantics were not adequately represented in the observed outputs for detailed evidential examination. Within the conditions evaluated here, the results support considering EFM3D primarily for contextual documentation and scene review rather than as a replacement for validated high-precision capture methods.

More broadly, the findings reinforce the need to evaluate foundation models in the operational settings for which they are intended. As these models move beyond general benchmarks into specialised and safety-critical domains, application-specific evaluation will be essential for distinguishing technical capability from practical readiness. The proposed forensic evaluation framework provides a structured basis for assessing future egocentric foundation models within crime scene reconstruction and for designing comparative evaluations with ground-truth geometry, evidence-level annotations, and practitioner assessment.

\bibliographystyle{spiebib}
\bibliography{Reference}
\end{document}